\documentclass[11pt,a4paper]{article}
\usepackage[hyperref]{acl2020}
\usepackage{times}
\usepackage{latexsym}

\usepackage{microtype}

\aclfinalcopy 

\usepackage{tabularx}
\usepackage{import}
\usepackage{graphicx}
\usepackage{amsmath}
\usepackage{makecell}
\usepackage{array}
\newcolumntype{?}{!{\vrule width 1.5pt}}

\usepackage{import}
\usepackage{color} 
\usepackage[multiple]{footmisc}
\usepackage[normalem]{ulem} 
\usepackage{enumitem}
\usepackage{booktabs} 

\definecolor{darkGreen}{rgb}{0,0.6,0}

\newcommand{\jk}[1]{\textcolor{blue}{[JK - #1]}}
\newcommand{\nh}[1]{\textcolor{violet}{[NH - #1]}}

\newcommand{\usage}[1]{{\color{red} {#1}}} 
\newcommand{\cut}[1]{\textcolor{red}{\sout{#1}}}
\newcommand{\hide}[1]{}

\usepackage[title]{appendix}

\title{Stimulating Creativity with FunLines:\\A Case Study of Humor Generation in Headlines} 

\author{Nabil Hossain$^*$, John Krumm$^\dagger$, Tanvir Sajed$^\ddagger$ \and Henry Kautz$^*$
\\
\\
$^*$Department of Computer Science, University of Rochester \\ 
$^\dagger$Microsoft Research AI, Microsoft Corporation, Redmond, WA \\
$^\ddagger$Department of Computing Science, University of Alberta \\ 
\small{\tt \{nhossain,kautz\}@cs.rochester.edu, jckrumm@microsoft.com, tsajed@ualberta.ca}
}

\date{}

\begin{document}
\maketitle


\begin{abstract}
Building datasets of creative text, such as humor, is quite challenging. We introduce \textbf{FunLines}, a competitive game where players edit news headlines to make them funny, and where they rate the funniness of headlines edited by others. FunLines makes the humor generation process fun, interactive, collaborative, rewarding and educational, keeping players engaged and providing humor data at a very low cost compared to traditional crowdsourcing approaches. FunLines offers useful performance feedback, assisting players in getting better over time at generating and assessing humor, as our analysis shows. This helps to further increase the quality of the generated dataset. We show the effectiveness of this data by training humor classification models that outperform a previous benchmark, and we release this dataset to the public.

\end{abstract}

\begin{figure*}
    \centering
    \begin{tabular}{c}
      \resizebox{0.9\linewidth}{!}{\fbox{\includegraphics{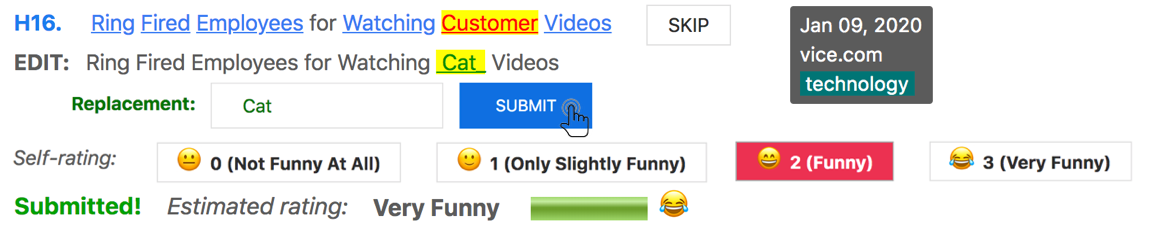}}} \\     (a) The Headline Editing Task. \\ 
      \resizebox{0.97\linewidth}{!}{\fbox{\includegraphics{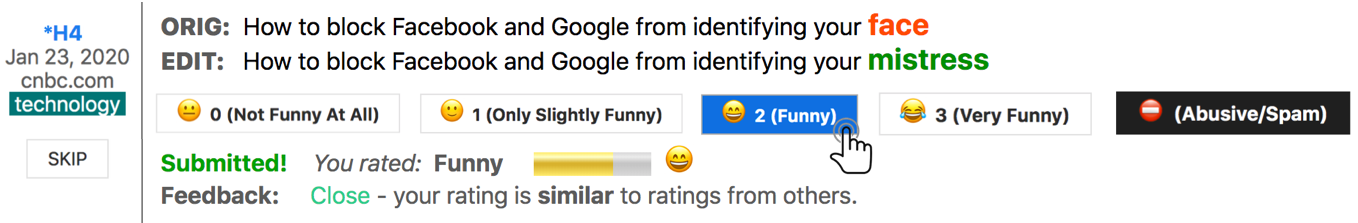}}} \\     (b) The Headline Rating Task.
    
    \end{tabular}
    \caption{Screenshots of the FunLines humor tasks.}
    \label{fig:tasks}
\end{figure*}

\section{Introduction}

While some data for machine learning tasks, like image object detection, is relatively easy to annotate, generating data that depends on human creativity is quite difficult.
Unlike many objective tasks, creativity is much less constrained, people do not always agree on the quality of creative output, and measuring creativity often requires more effort.
One such creative endeavor is humor, which is subjective and hard to define, making consensus difficult. Factors such as human preferences, mood, world knowledge, and complex interrelationships play important roles in determining what is funny, making it challenging to generate humor datasets. 

One approach to crowdsourcing creative data is to hire workers, as we did for humorous headlines in~\citet{hossain-etal-2019-president}. An alternative is to make a game for gathering the data. Games are especially suitable for gathering creative output because it is enjoyable to both generate and rate creative artifacts, particularly humor. The effectiveness of games for labeling data has been shown in previous work~\cite{von2004labeling,von2008designing}.

In this paper, we introduce Funlines\footnote{\href{https://funlines.co}{\tt{https://funlines.co}}
\\FunLines demo video:  \href{https://youtu.be/5OXJMxDBaLY}{https://youtu.be/5OXJMxDBaLY} }, an online  game for generating funny news headlines for humor research. 
We explore and evaluate this fun, competitive way of motivating people to contribute creative text, addressing some of the special challenges of generating humor data mentioned above.

Generating a dataset of creative artifacts for machine learning requires ratings of the artifacts. This has been done for sites like Reddit (\texttt{reddit.com}) for jokes and Photofeeler     
(\texttt{photofeeler.com}) for photos, both of which allow casual users to upvote/downvote creative content from others. The Hafez research project lets users rate machine generated poetry~\cite{ghazvininejad-etal-2017-hafez}. In FunLines, one of the two main tasks for players is to rate the funniness of headlines that have been edited  by other players. Humor is inherently social, and the headlines' collective exposure to, and rating from, humans preserves this important feature.

The other main task in FunLines is to edit regular headlines to make them funny. To simplify machine learning and humor analysis, we want funny headlines that are highly constrained. In FunLines, players start with a regular headline from a news source and they change a single word or entity to make it funny, giving data that is particularly suitable for understanding the tipping point between serious and humorous. This contrasts with Unfun.me, which is also a game for generating pairs of funny and unfunny headlines~\cite{west2019reverse}. With Unfun.me, players start with a funny headline and make it serious, where their edits are unconstrained but encouraged to be minimal. Because it starts with regular headlines, FunLines has an enormous amount of raw source material. Also related to FunLines' editing tasks are projects that help users create humor, like HumorTools~\cite{chilton2016humortools} and Libitum~\cite{hossain2017filling}.

By deploying FunLines, we learned how players behaved, and we responded with modifications to encourage players to provide creative, high quality data. We show that over time, players made significant improvements in both the quality of their humor and consistency of their ratings. 
We also show that the resulting data is effective for training humor detection systems. In fact, FunLines uses one of our humor detectors to give instant feedback when players edit a headline. We compare the FunLines data to {\bf Humicroedit}, which is a humorous headlines dataset we collected previously from crowd workers~\cite{hossain-etal-2019-president}. FunLines data is less expensive, higher quality, and leads to improved automated humor detection. Overall, we show that a competitive game is effective for gathering data for a human creativity task based on an experiment with a large number of users.

\section{The FunLines Humor Game}
FunLines is designed to collect a large volume of rated, humorous headlines. It makes the humor generation process fun, interactive, competitive, rewarding and educational, keeping players engaged.

In FunLines, players can attempt two tasks: (i) edit regular news headlines to make them funny, and (ii) rate the funniness of headlines edited by other players. They receive feedback for these actions, which helps them get better at the game.
Players are ranked in our performance-based leaderboards, which offer prize money. 
We now describe the various aspects of the game that make it work.

\subsection{Editing Headlines}
Similar to our previous work~\cite{hossain-etal-2019-president}, we restrict the headline editing task to 
the substitution of a noun, verb or entity in the headline with a single word. This constraint enables focused analysis on humor triggered by atomic changes in text.

Shown in Figure~\ref{fig:tasks}(a), the headline editing interface highlights the headline's replaceable words in \textcolor{blue}{\underline{blue}}. It allows a single word substitute to be submitted by the player.
To help put the headline in context, we provide information such as the news source, category and date of publication, including a link to the article (\textcolor{blue}{H16} in Fig.~\ref{fig:tasks}(a)). Upon submitting the edited headline, the player receives an estimated funniness score from a humor classifier built into FunLines (see Section~\ref{sec:experiments}).

\subsection{Rating Headlines}
As in~\citet{hossain-etal-2019-president}, players rate a headline's funniness on a 4 point scale, as shown in Fig.~\ref{fig:tasks}(b):

\vspace{1.2pt}
\begin{tabular}{ll}
{\bf 0} - Not funny at all \includegraphics[height=0.8em]{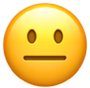}   &  {\bf 2} - Funny \includegraphics[height=0.8em]{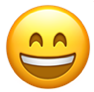}  \\
{\bf 1} - Only slightly funny \includegraphics[height=0.8em]{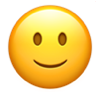} &   {\bf 3} - Very funny \includegraphics[height=0.8em]{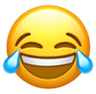}
\end{tabular}
\vspace{0.3pt}

\noindent 
We provide instant rating feedback to players when they rate a headline that already has a consensus rating from other players. 
This feedback includes whether the player's rating is reasonably close, higher, or lower compared to ratings from others. 




Each edited headline is made available in the game until it has {\bf five} funniness ratings, which constitutes a {\bf fully rated} edited headline.

\subsection{Competition Setting}
We designed FunLines as a humor competition to motivate players to perform well. Players are scored based on their editing and rating, and they are ranked on the game's leaderboard page.

\subsubsection{Leaderboard}

We maintain two performance-based leaderboards: 
\begin{enumerate}[leftmargin=*]
\itemsep-0.2em 
    \item \textbf{Top points scorers:} This ranks players by points scored based on volume and quality of editing and rating as well as keeping a good balance between the volume of editing and rating.
    \item \textbf{Highest average ratings:} This ranks players who received the highest mean funniness ratings for their edited headlines. 
\end{enumerate}
Players who rank in the top 10 positions in any of the two leaderboards at the end of the 5-week long contest receive prize money between US\$ 5-100.

The leaderboard page also lists the recent top 10 funny headlines to show players examples of 
successful edits so that they can adjust their editing style, and to encourage them to make it to this list.
\hide{Need to mention budget is USD 1000. And mention that although leaderboard 2 ignores a player's ratings, by rating well the player can be considered for prize money from multiple leaderboards. And leaderboard 2 is also there to bias the data collection towards funny headlines, i.e., those players who solely compete for leaderboard 2 are greatly incentivized to make each of their edited headlines high quality funny.}

\subsubsection{Qualification Requirements}
Our competition requires participants to edit between 50-150 headlines and to rate between 200-500 headlines to be considered for prize money. The upper limits prevent players from running up their scores simply based on volume rather than quality. The lower limits help us collect more data, and since players get better at the game over time (details in Section~\ref{sec:behavior}), the quality of data improves with a larger volume of work from each player.



\subsubsection{Scoring}
Players receive humor points when they edit or rate headlines. We do not reveal the precise scoring formulae to the players to discourage them from ``gaming'' the game.
The three scoring factors are:


\noindent \textbf{Editing Points:} This score component exponentially increases with a headline's funniness ratings, encouraging players to make each edited headline as funny as possible.

\noindent \textbf{Rating Points:} This is a function of the difference between the player's ratings and ratings from other players for the headline. The lower the difference, the higher the points earned, and a large difference gives the player negative points.
This incentivizes players to be objective, rating headlines based on how others (\emph{e.g.,} a crowd) would rate them instead of letting their own biases influence their decisions. 

\noindent \textbf{Task Balancing Points:} 
Players are rewarded for maintaining a good balance between the numbers of edits and ratings. These points are maximized if the ratio of total ratings to total edits is between 3-10.
This discourages players from ignoring one of the tasks, 
helping to keep our dataset balanced. 

\subsection{Protecting Against Abusive Behavior}
During pilot tests, we noticed certain abusive player behavior, which we minimized, as described below: 

\noindent {\bf Abusive Edits:} We prevent players from submitting slang words, crude sexual references, bathroom jokes and other cheap forms of humor using blacklists. Also, players are encouraged to flag an edited headline if it demonstrates abusive behavior, 
and such headlines are removed from the rating pool, depriving their editors of potential points.

\noindent {\bf Abusive Ratings:} We 
forced a time delay 
to prevent players from rapidly rating headlines without reading them. 
Attempts at lowering others' points by consistently assigning low ratings cause players to accumulate large negative points as these ratings mostly disagree with ratings from other players.

Players who repeatedly show abusive behavior are warned, and, in some cases, suspended in order to maintain a healthy competition environment. 

\subsection{Performance Feedback} \label{sec:PerformanceFeedback}
FunLines gives players feedback on their performance so that they can improve their play:

\noindent \textbf{Editing:} Players see their top 5 most funny edited headlines and their 10 most recent edited headlines and the corresponding ratings for all. This helps them monitor how their edited headlines appeal to other players, and to adjust their editing style. They also see which of their edits are marked as abusive. 

\noindent \textbf{Rating:} Players see the histogram of their rating selections and the percentage of their ratings that are significantly above or below the ratings of others who rated the same headlines. This helps them spot and rectify errors such as frequently overestimating or underestimating ratings, a common user behavior we saw initially in the competition. Players also see the 10 most recent headlines they rated and the ratings they assigned. 

Players are advised on how many more edits or ratings to do to optimize their task balancing score.



\subsection{Ordering Headlines for Rating}
The ordering of headlines displayed for rating greatly influences which headlines receive more ratings. Using their sampling weights, we re-compute the display order 
of headlines for rating every few minutes. Headlines are assigned higher sampling weights if they: (i) are from players with high volumes of edited headlines, (ii) are from players who received very few ratings, (iii) are recently submitted, or (iv) have received more 
ratings.

Ratings are diversified by limiting each player from rating more than 10 headlines of another player.
These design decisions make the game fair, engaging, and rewarding to all players. 

\subsection{Popular Headlines and Player Flexibility}
FunLines provides headlines from 5 diverse topics: politics, world news, entertainment, technology and sports. It daily adds about 300 trending English headlines from major news outlets posted to Reddit.

FunLines gives players the flexibility to choose which headlines to edit or rate, a freedom that the crowd workers in~\citet{hossain-etal-2019-president} did not have. 
Not all headlines are easy to edit or judge for humor, 
and by providing a large pool of headlines, FunLines allows players be strategic and selective. For example, they can focus only on headlines they appreciate and understand,  
and they can use the SKIP button to permanently ignore headlines that are confusing or simply too difficult to edit or rate. 

Our editable headlines are sorted by most recently published first, making it easier for players to generate humor with stories that are fresh in their minds. 
FunLines also offers players the option of choosing to work on headlines by their preferred news categories. Overall, these features help tackle human preferences and knowledge related challenges associated with humor understanding.

\section{Gathering Data}
In this section, we describe our player recruiting strategies, our gathered dataset, and its comparison against Humicroedit~\cite{hossain-etal-2019-president}.

\subsection{Attracting Players}
To jump start data collection, we hired players from Amazon Mechanical Turk (AMT). Our competition budget was US\$ 1,000, and since the total prize money was US\$ 560, we spent the remaining US\$ 440 to hire 100 US turkers. These turkers were intended to seed the game, and many of them continued playing after their paid work was finished. 


We sought additional players by advertising FunLines in our networks, making a TV news appearance, and posting on relevant social media pages. 


Overall, we had 290 players, out of whom 
204 completed at least 20 edits and ratings,
89 met the competition qualification requirements, 
and 33 completed the maximum 150 edits and 500 ratings.


\begin{table}
    \centering
    \resizebox{\columnwidth}{!}{
    \begin{tabular}{|l|l|c|}
         \textbf{Original Headline} (replaced word in \textcolor{red}{\textbf{bold}}) & \textbf{Edit} & \textbf{RT} \\ \hline
         Sanders says he has more \textcolor{red}{\textbf{donors}} than Trump & \textcolor{darkGreen}{hair} & 3.0 \\ \hline
         `What is the green \textcolor{red}{\textbf{new}} deal?' & \textcolor{darkGreen}{ham} & 2.4 \\ \hline
         Japan begins controversial \textcolor{red}{\textbf{Taiji}} dolphin hunt & \textcolor{darkGreen}{ninja} & 2.0 \\ \hline 
         Bolton confirms he's willing to \textcolor{red}{\textbf{testify}} & \textcolor{darkGreen}{lie} & 1.6 \\ \hline
         A\$AP Rocky found guilty of \textcolor{red}{\textbf{assault}} in Sweden & \textcolor{darkGreen}{singing} & 1.0 \\ \hline
         \textcolor{red}{\textbf{Netherlands}} to drop 'Holland' as nickname & \textcolor{darkGreen}{Gangster} & 0.4  \\ \hline
         The useful \textcolor{red}{\textbf{idiot}} from Louisiana & \textcolor{darkGreen}{Louis} & 0.0 \\ \hline
         K-pop star Sulli found dead aged 25 & \emph{skipped} & -- \\ \hline
    \end{tabular}}
    \caption{Headlines in FunLines with ratings.}
    \label{tab:examples}
\end{table}

\subsection{Dataset Quality}
\begin{table}
    \centering
    \resizebox{0.9\columnwidth}{!}{
        \begin{tabular}{l|c|c}
            \textbf{Metric}         & \textbf{Humicroedit}   & \textbf{FunLines} \\ \hline
            Size                    & 15,095        & 8,248 \\ \hline
            Mean funniness          & 0.94          & 1.26 \\ \hline
            Cost per datum          & 29.8c         & 12.1c \\ \hline
            Agreement $\alpha$      & 0.20          & 0.25 \\ \hline
            Unique words used       & 41.2\%        & 53.4\% \\ \hline
            No. of editors          & 73            & 214 \\ \hline
            No. of raters           & 131           & 246 \\ \hline
        \end{tabular}}
    \caption{Humicroedit and FunLines data comparison.}
    \label{tab:dataset_comparison}
\end{table}



Here we examine the FunLines dataset\footnote{Dataset: \href{www.cs.rochester.edu/u/nhossain/funlines.html}{cs.rochester.edu/u/nhossain/funlines.html}} and we compare it to Humicroedit~\cite{hossain-etal-2019-president},
the same type of headline data we previously obtained using only AMT workers and without a game. 
Table~\ref{tab:examples} shows sample headlines in FunLines and Table~\ref{tab:dataset_comparison} shows several quality measures.

In total, we received 13,063 edited headlines and 46,359 headline ratings, leading to 8,248 fully rated headlines and only 55 abusive headlines. On average, players took about 25 seconds per edit and 5 seconds per rating, implying that they collectively spent about 168 hours playing the game.
Given that our budget was US\$ 1,000, this makes the hourly participation rate only US\$ 5.95. The cost of each fully rated headline in FunLines is 12.1 cents vs 29.8 cents in Humicroedit, a nearly 60\% cost reduction. The mean funniness rating for Humicroedit is 0.94 vs. 1.27 in FunLines, meaning that our competition based approach enabled us to achieve a 35\% boost in funniness on average. Funnier headlines are hard to obtain, which makes our FunLines dataset more valuable for machine learning. 

The annotator agreement score in FunLines, measured by Krippendorf's $\alpha$~\cite{krippendorff1970estimating}, is 0.25 vs. 0.20 in Humicroedit, implying higher agreement without spending funds for qualifying players, which we did for Humicroedit. 
Instead, the game's real-time feedback helps people learn how to be better editors and raters. 
53.4\% of the words used in FunLines are unique vs. 41.2\% in Humicroedit, indicating more diversity of edits. 

All these factors show that our competition based approach to humor data collection is a better option than simply hiring crowd workers who are not incentivized and rewarded based on performance.

Finally, our data size could be increased by raising the editing and rating caps, running the competition longer, reaching out to more people, or being more efficient at rating partially rated headlines.


\section{Analysis of Player Behavior}
We examine how players adapted to the FunLines competition and the headlines they attempted.



\subsection{Player Improvement over Time}
\label{sec:behavior}

\begin{figure*}
    \centering
    \resizebox{\linewidth}{!}{
    \begin{tabular}{ccc}
    \includegraphics{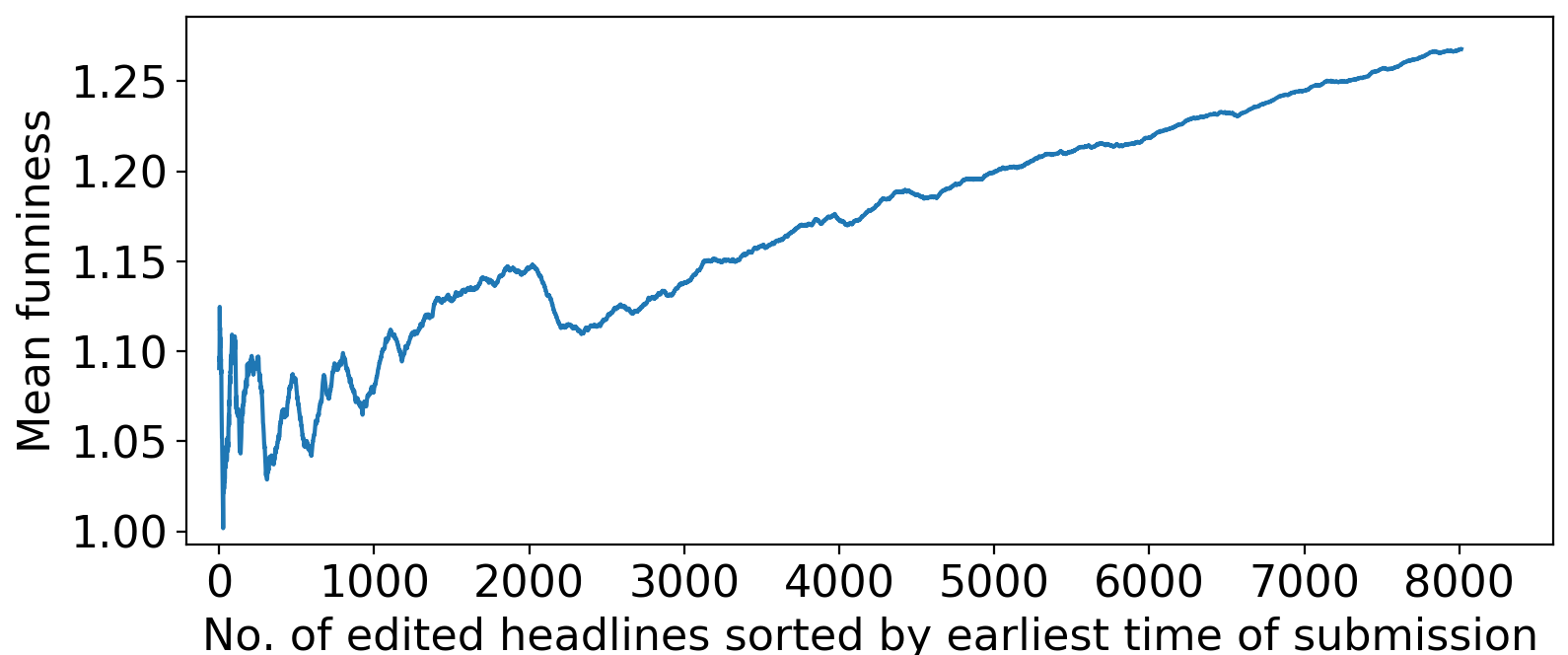} &
    \includegraphics{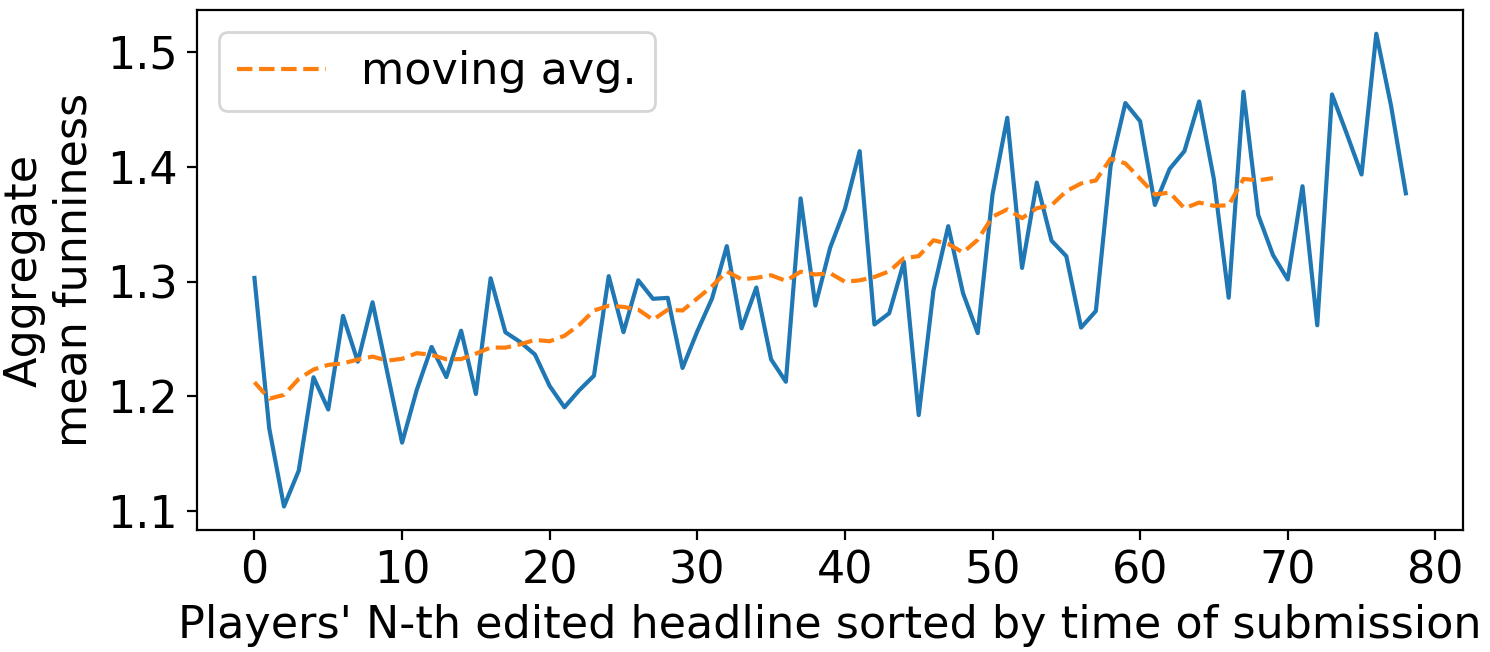} &
    \includegraphics{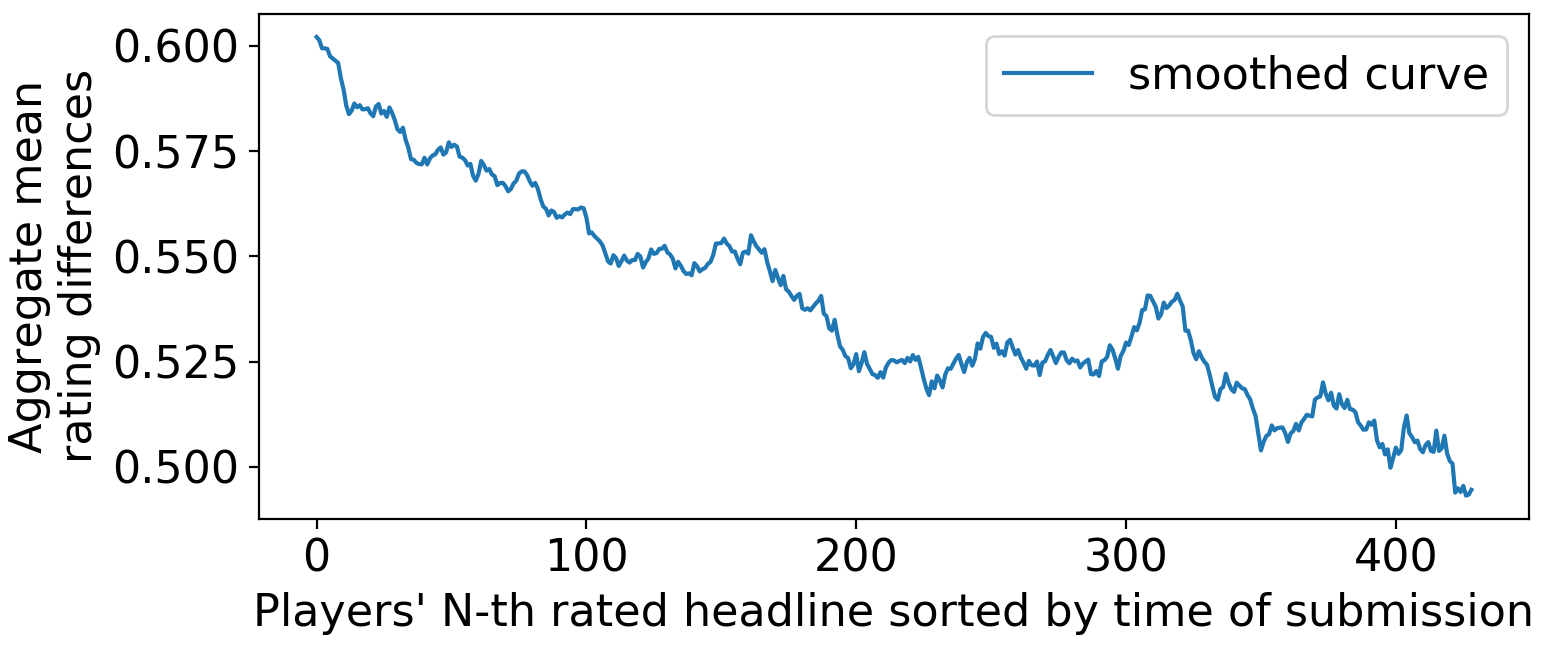} \\ 
    \end{tabular}}
    \vspace{3pt}
\phantom{x}\hspace{20pt} (a) Dataset's funniness. \hspace{30pt} (b) Players' editing quality.  \hspace{20pt} (c) Players' rating differences.
    \caption{Players get better at editing and rating headlines over time, helping to increase funniness in the dataset.}
    \label{fig:temporal}
\end{figure*}

Figure~\ref{fig:temporal}(a) shows that over time, the mean funniness of fully rated headlines steadily increases, indicating that players are getting better at making funnier edited headlines as the competition proceeds. In addition, Figure~\ref{fig:temporal}(b) shows that as players edit more headlines, they generally get better at making funnier edits. 
Finally, Figure~\ref{fig:temporal}(c) shows that as players rate more headlines, 
the difference of their ratings from the mean ratings of others for the same headline decreases, indicating that players become more consistent at rating over time. 

Overall, Figure~\ref{fig:temporal} demonstrates the educational aspect of FunLines, suggesting that by playing the game, people can learn to be better at generating humor and agreeing on its ratings. 
These results emphasize the advantages of this competitive and collaborative approach in helping us collect quality humor data, mirroring the social nature of humor. 

\subsection{Choosing Headline Categories}
We analyze whether players were strategic about choosing headlines to edit. Table~\ref{tab:news_categories} shows the proportions of the five headline categories in the headlines supplied by FunLines (H), the proportions of headlines from each category that were edited (E), the proportions of each category in the fully rated FunLines dataset (FR), and the mean funniness ratings of headlines from each category.


The two classes with the highest mean funniness ratings are politics and world news. The E (edited) column shows that players are ignoring about 30\% of headlines from these categories, while they are almost exhausting the headlines from the other three categories.
Not all headlines can be easily made funny, and attempting almost all the headlines from these three categories is likely hurting players' funniness scores. Besides, politics and world news are very popular topics, whereas the others are somewhat niche categories with smaller audiences, and thus the raters are perhaps not familiar with the entities and events mentioned in their articles, and thus might not be ``getting'' the jokes. 


\begin{table}
    \centering
    \resizebox{0.97\columnwidth}{!}{
    \begin{tabular}{l|c|c|c|c}
        \textbf{Category}        & \% \textbf{H}  \hide{& \% ED}    & \% \textbf{E} & \% \textbf{FR}  & \textbf{Mean RT} \\ \hline
        Politics        & 70.0   \hide{& 59.8}     & 71.5 & 56.4     & 1.32 \\ \hline 
        World news      & 22.3   \hide{& 22.5}     & 71.0 & 22.4     & 1.29 \\ \hline 
        Technology      & 4.6    \hide{& 9.6}      & 92.9 & 11.0     & 1.22 \\ \hline 
        Sports          & 1.1    \hide{& 2.8}      & 96.7 & 3.4      & 1.14 \\ \hline 
        Entertainment   & 2.0    \hide{& 5.3}      & 95.2 & 6.8      & 0.96 \\ \hline 
    \end{tabular}}
    \caption{FunLines headline categories with ratings.}
    \label{tab:news_categories}
\end{table}






\section{Detecting Humorous Headlines}


We investigate if the FunLines dataset of 8,248 fully rated headlines is suitable for humor detection.
\subsection{Instant Editing Feedback}
\label{sec:experiments}
\begin{figure}
    \centering
    \includegraphics[width=0.9\columnwidth]{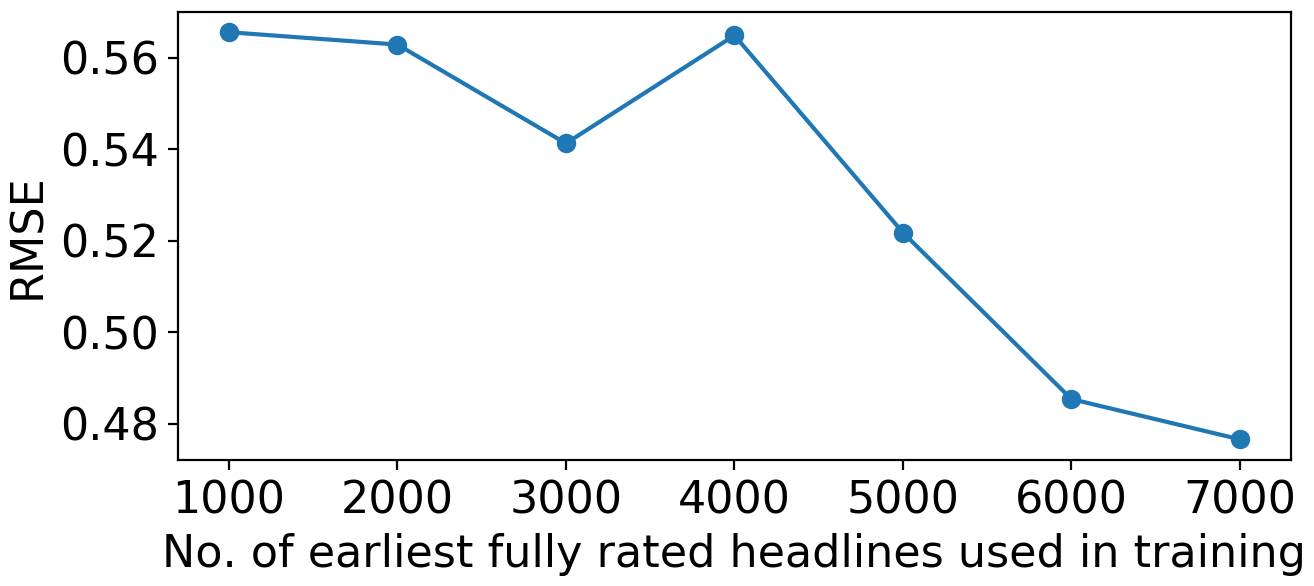}
    \caption{Results of re-training BERT funniness regression as more data becomes available in FunLines.}
    \label{fig:regression}
\end{figure}

Whenever a player submits an edited headline, FunLines uses a humor detection system to provide instant funniness feedback.
This is a fine-tuned BERT next sentence prediction model~\cite{devlin-etal-2019-bert} in the regression setting that uses the input:
\begin{center}
$<$original headline$>$ \texttt{[SEP]} $<$edited headline$>$ 
\end{center}
We re-train this model as blocks of new data becomes available. 
We start by training on the first 1,000 fully rated headlines, and we test our model against the next 1,000. Then we train on the first 2,000 and test on the next 1,000, and so on. The results, shown in Figure~\ref{fig:regression}, suggest that the model gets increasingly accurate in its funniness estimations over time as more data becomes available, and thus its feedback to the user improves over time. We attribute this to both the increased volume of data and the increased quality and consistency of edited and rated headlines, as illustrated in Figure~\ref{fig:temporal}. 

\subsection{Improving Classification for Humicroedit}
We explore whether using the FunLines data can improve binary humor classification in Humicroedit~\cite{hossain-etal-2019-president}. 
We augment the training data of Humicroedit with the FunLines data, and we evaluate classification performance with and without this dataset augmentation. 

Humicroedit was trained on the top and bottom $X\%$ funny and unfunny headlines in its dataset. Each $X\%$ gives an upper and lower funniness threshold, and we use these same thresholds (MaxUF and MinF in Table~\ref{tab:naacl_classification}) to select the augmentation data from FunLines. We trimmed the FunLines data so there was balanced augmentation for funny and unfunny.


For a fair comparison with~\citet{hossain-etal-2019-president}, 
we ignore the original headlines and use only the
edited versions of headlines as classifier input.
We fine-tune the BERT sentence classification model on both the original and augmented Humicroedit dataset. In  Table~\ref{tab:naacl_classification}, we show the funniness thresholds for each subset $X$, the benchmark results in~\citet{hossain-etal-2019-president} obtained using LSTM with GloVe word vectors, and we report our new results. 

These results suggest that while BERT trained on Humicroedit alone outperforms LSTM, the FunLines data helps further BERT's classification accuracy (BERT Aug.) by up to 2\% for each of the sub-datasets. This is good improvement given that we are only augmenting Humicroedit with part of the FunLines dataset and the 0-3 funniness scale between the two datasets are not calibrated. 
Further, in Humicroedit, each original headline was edited three times, so there is overlap in its training and test sets, which makes the task a bit easier. In FunLines, headlines were mostly edited only once. The timeline between the two datasets was distinct, minimizing the possibility of repeated headlines between the two.

\begin{table}
    \centering
    \resizebox{\columnwidth}{!}{
    \begin{tabular}{c|c|c|c|c|c}
        X   & MaxUF     & MinF      & LSTM      & BERT      & BERT Aug. \\ \hline
        10  & 0.2       & 1.8       & 68.54     & 76.48     & 77.32 \\ \hline
        20  & 0.4       & 1.4       & 67.21     & 73.76     & 75.66 \\ \hline
        30  & 0.6       & 1.2       & 66.11     & 69.37     & 70.75 \\ \hline
        40  & 0.8       & 1.0       & 64.07     & 66.38     & 68.00 \\ \hline
        50  & 0.8       & 0.8       & 60.6      & 63.93     & 64.40 \\ \hline
        
    \end{tabular}}
    \caption{Humicroedit classification accuracy with and without using augmented training data from FunLines. }
    \label{tab:naacl_classification}
\end{table}

Our experiments used the BERT base model with 
a learning rate of $1e^{-4}$, max seq length of 64, batch size of 8, and we trained models for up to 3 epochs. 


\section{Conclusion and Future Work}

FunLines is an online game for generating funny headlines. While creative data can be difficult to obtain, FunLines makes it easier by taking advantage of the inherent fun of creativity and competition. We described the game, including a rich set of feedback for players to assess their own performance along with controls and incentives for them to create funny headlines. Our deployment attracted 290 players for a total cost of US\$ 1,000.
Compared to our earlier work for gathering the same type of data from only turkers, FunLines produced funnier headlines with better rating agreement and at nearly 60\% lower cost per headline. We showed how players' performance improves over time, both in terms of their headline quality and rating consistency. We showed how the FunLines data is effective for training machine learning models to detect and to rate humorous headlines. We have shared this data in an ongoing SemEval task for humor recognition~\cite{hossain2020semeval}\footnote{\href{https://competitions.codalab.org/competitions/20970}{https://competitions.codalab.org/competitions/20970}}.

Having built models for detecting humor, we see this data as the foundation of the automatic \emph{creation} of humorous headlines in a generate-and-test approach. More generally, FunLines is a prototype for gathering datasets of creative artifacts from people that is engaging, interactive, competitive, rewarding, educational and inexpensive.


\bibliography{humor}
\bibliographystyle{acl_natbib}


\end{document}